\documentclass{article}

\usepackage[preprint]{neurips_2025}

\usepackage[utf8]{inputenc} 
\usepackage[T1]{fontenc}    
\usepackage[pagebackref=true,colorlinks,citecolor=brown]{hyperref}       
\usepackage{url}            
\usepackage{booktabs}       
\usepackage{amsfonts}       
\usepackage{nicefrac}       
\usepackage{microtype}      
\usepackage{xcolor}         
\usepackage{xspace}
\usepackage{graphicx}
\usepackage{multirow}
\usepackage{multicol}
\usepackage{bm}
\usepackage{pifont}
\usepackage{amsmath}
\usepackage{todonotes}
\usepackage{longtable}
\usepackage{arydshln}
\usepackage{makecell}
\usepackage{amssymb}

\usepackage{wrapfig}
\usepackage{lipsum}
\usepackage{caption}

\definecolor{darkred}{RGB}{196,0,0}

\newcommand{\Model}{Mirage\xspace}

\makeatletter
\def\@fnsymbol#1{\ensuremath{\ifcase#1\or \dagger\or \ddagger\or
\mathsection\or \mathparagraph\or \|\or **\or \dagger\dagger
\or \ddagger\ddagger \else\@ctrerr\fi}}
\newcommand{\printfnsymbol}[1]{%
  \textsuperscript{\@fnsymbol{#1}}%
}
\makeatother

\title{Machine Mental Imagery: Empower Multimodal Reasoning with Latent Visual Tokens}

\newcommand*{\affmark}[1][*]{\textsuperscript{#1}}

\author{
Zeyuan Yang\affmark[1]\thanks{Equal contribution}~~~ 
Xueyang Yu\affmark[1]\footnotemark[1]~~~
Delin Chen\affmark[1]~~~
Maohao Shen\affmark[2]~~~
Chuang Gan\affmark[1]~~~
\\ \\
$^1$University of Massachusetts, Amherst~~~~~~
$^2$Massachusetts Institute of Technology \\ \\
\begin{tabular}{@{}ll@{}}
\textbf{Project Page:} & \url{https://vlm-mirage.github.io} \\
\textbf{Code:} & \url{https://github.com/UMass-Embodied-AGI/Mirage} \\
\end{tabular}
}

\begin{document}

\maketitle

\begin{abstract}
Vision-language models (VLMs) excel at multimodal understanding, yet their text-only decoding forces them to verbalize visual reasoning, limiting performance on tasks that demand visual imagination.
Recent attempts train VLMs to render explicit images, but the heavy image-generation pre-training often hinders the reasoning ability.
Inspired by the way humans reason with mental imagery—the internal construction and manipulation of visual cues—we investigate whether VLMs can reason through interleaved multimodal trajectories without producing explicit images.
To this end, we present a Machine Mental Imagery framework, dubbed as \textbf{\Model}, which augments VLM decoding with latent visual tokens alongside ordinary text.
Concretely, whenever the model chooses to ``think visually'', it recasts its hidden states as next tokens, thereby continuing a multimodal trajectory without generating pixel-level images.
Begin by supervising the latent tokens through distillation from ground-truth image embeddings, we then switch to text-only supervision to make the latent trajectory align tightly with the task objective. A subsequent reinforcement learning stage further enhances the multimodal reasoning capability.
Experiments on diverse benchmarks demonstrate that \Model unlocks stronger multimodal reasoning without explicit image generation. 
\end{abstract}

\section{Introduction}

Vision–language models (VLMs) jointly encode images and text and attain impressive results on visual-understanding benchmarks through text-only decoding~\citep{wang2024qwen2}. Techniques such as chain-of-thought prompting and reinforcement-learning fine-tuning can lengthen these textual reasoning traces and yield extra gains.
Nonetheless, VLMs still stumble on multimodal reasoning tasks such as spatial reasoning, which demand more than passive perception; they require a coherent understanding and manipulation of visual elements.

\begin{figure}[t]
\centering
\includegraphics[width=1\linewidth]{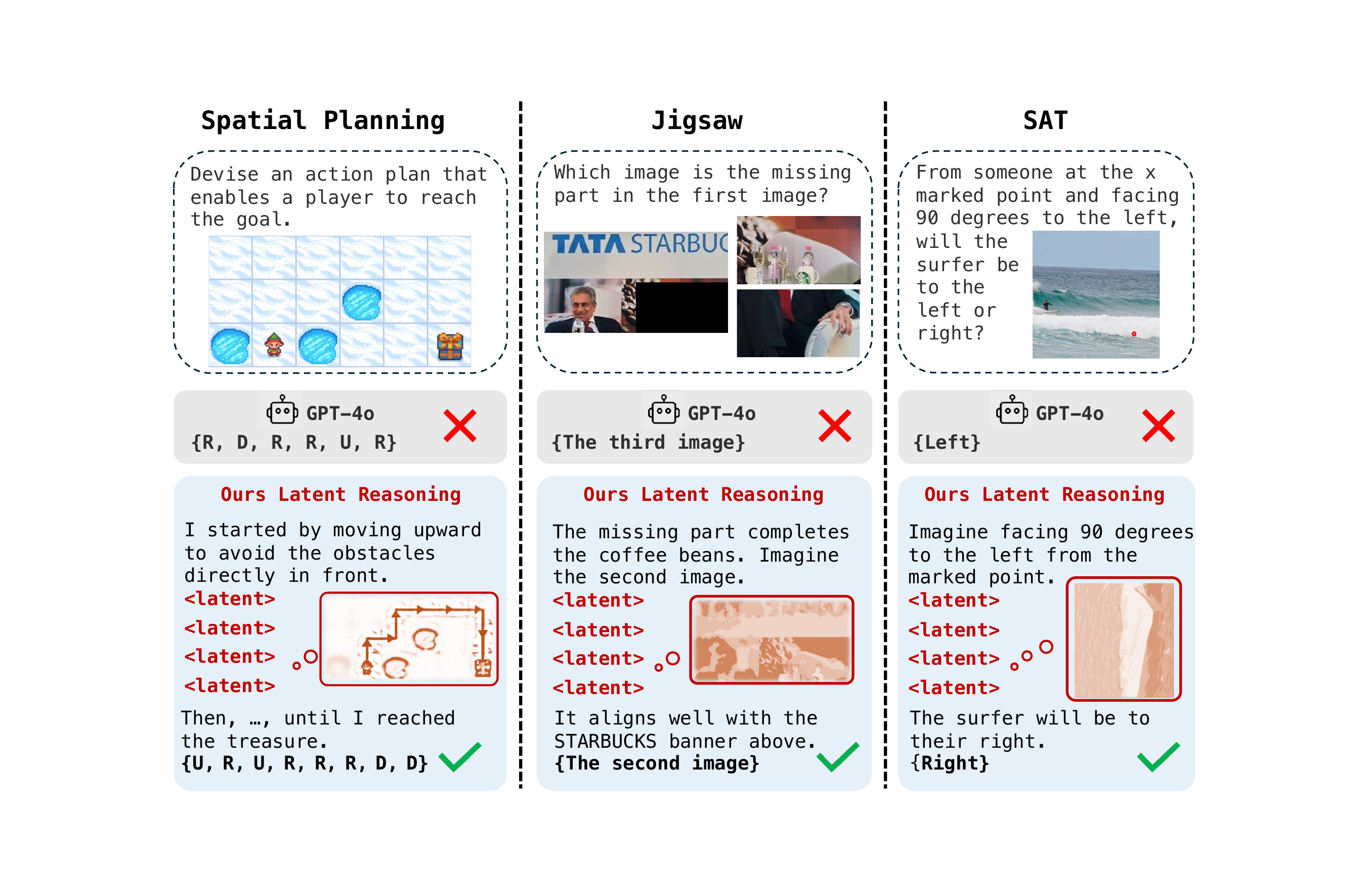}
\caption{\textbf{Multimodal Reasoning Examples.} \Model interleaves \textcolor{darkred}{\textbf{latent visual tokens}}, which represent compact imagery visual features, with explicit text tokens to solve diverse spatial reasoning multimodal tasks, boosting the reasoning performance without the full pixel-level image generation.}
\label{fig:teaser}
\end{figure}

Consider the jigsaw puzzle in Fig.~\ref{fig:teaser}.
Instead of textualizing every candidate piece, people picture how the two fragments might align and decide on the correct match.
This reasoning unfolds in a native multimodal fashion, not through language alone.
Recent studies~\citep{team2024chameleon,tong2024metamorph,chern2024anole,chen2025blip3} have pre-trained VLMs for large-scale image generation so a single model can produce both words and pictures.
Yet the cognitive demands of logical reasoning differ sharply from the task of synthesizing pixels, and asking one model to master both goals often degrades its reasoning quality~\citep{wang2025autoregressive}.
In addition, the image decoders cannot produce interleaved trajectories pertinent to input images.
Consequently, fully exploiting the dormant multimodal reasoning capacity of VLMs remains an open challenge.

According to imagery theory, humans do not summon photorealistic pictures while thinking. We instead construct and manipulate mental images, simplified sketches that capture only task-relevant information, a process known as \textbf{mental imagery}~\citep{shepard1971mental, farah1985psychophysical, kosslyn1996image}. In the jigsaw example, we examine fragment contours to decide whether two pieces fit. Likewise, when searching for misplaced keys, we recall the outline of the shelf edge rather than the full room. Inspired by this behavior, we ask whether VLMs can reason directly in their latent visual embedding space, weaving compact visual embeddings into the text stream and dispensing with the need for explicit image generation.

To this end, we present \textbf{\Model}, a decoding mechanism that interleaves latent visual representations among text tokens.
Prior studies have shown that LLMs can reason directly within the latent space. Building upon this insight, in our \Model framework, when the model chooses to reason visually by producing a special token, it then reuses its current hidden state as a compact visual embedding and appends it to the context, skipping the language projection. These internal embeddings furnish focused visual cues for later reasoning steps. As illustrated in Fig.~\ref{fig:teaser}, \Model yields a chain-of-thought trajectory without any external image decoder.

\vspace{-2pt}

As illustrated in Fig.~\ref{fig:framework}, we adopt a two-stage fine-tuning paradigm to equip the model with interleaved reasoning. In the first stage, with annotated interleaving trajectories, we supervise both modalities: the model predicts the next word while reconstructing a compact latent visual vector obtained from compressed image embeddings. This dual objective anchors the latent tokens in the visual subspace and teaches the model to weave visual cues into its output.

The second stage removes direct supervision on the latent vectors and optimizes only the text tokens, letting the model treat its autoregressively generated latent embeddings as priors that guide subsequent word generation.
This relaxation yields a more flexible interleave reasoning trajectory without forcing the latent channel to match any predetermined embedding.
After these two stages, we apply reinforcement learning to further boost the reasoning performance.

Extensive experiments and superior performance across multiple benchmarks demonstrate that our proposed \Model significantly enhances the reasoning ability of VLMs compared with text-only decoding. More concretely, our contributions are threefold,

\begin{itemize}
    \item We introduce \Model, which enables VLMs to generate interleaved reasoning trajectories that mix latent visual tokens with ordinary text, without relying on external visual decoders.
    \item Our two-stage training paradigm empowers VLMs to produce stable yet flexible interleaved reasoning and shows that reinforcement learning can further boost performance.
    \item \Model achieves consistent gains across diverse multimodal reasoning benchmarks. Further analysis reveals that the latent tokens embody meaningful visual cues, underscoring the potential to unlock deeper multimodal reasoning capabilities in VLMs.
\end{itemize}

\section{Related Work}

\subsection{Multimodal Chain-of-Thought}
Chain-of-Thought (CoT) prompting was first shown to elicit step-by-step reasoning in LLMs by supplying a few worked examples that include intermediate rationales~\citep{feng2023towards,zhang2024chain, wei2023chainofthoughtpromptingelicitsreasoning}.
Recent extensions of CoT to multimodal settings embed visual evidence directly into the reasoning trajectory. ICoT ~\citep{zhang2024multimodalchainofthoughtreasoninglanguage} interleaves attention-selected image crops with text tokens, yielding significant VQA gains, while Visual CoT ~\citep{shao2024visualcotadvancingmultimodal} supplies 438 k bounding-box-grounded rationales to train VLMs that emit explicit visual tokens and improve spatial grounding.
Recent works~\citep{hu2024visual,zhou2024image,yang20253d,gao2024cantor, wu2025reinforcing, chern2025thinking,fang2025intention, cheng2025visual, su2025openthinkimg} further leverage external tools to supply visual cues that enrich multimodal CoT reasoning..

Recent works~\citep{chen2025blip3, wang2025autoregressive} like Chameleon~\citep{chameleonteam2025chameleonmixedmodalearlyfusionfoundation,chern2024anole} trains a unified token-based model that can emit arbitrary sequences of text and image tokens, but at the cost of large-scale pixel-level supervision and heavier decoding. MVoT~\citep{li2025imagine} further trains a unified model to directly produce image and text interleaving trajectories, but absent of reasoning thoughts.
In contrast, our \Model framework differs by emitting compact latent visual tokens rather than real image patches or pixels, avoiding heavy image generation while still allowing fully interleaved visual–text reasoning.

\subsection{Latent Reasoning in LLMs}

Much recent work has highlighted the importance of intermediate hidden representations in Large Language Models (LLMs)~\citep{biran2024hopping,yang2024large}. To better guide the latent reasoning process, several approaches introduce specialized tokens into the input sequence. ~\cite{wang2023guiding} incorporate discrete \texttt{<plan>} tokens to control reasoning stages, while ~\cite{goyal2023think} propose inserting a \texttt{<pause>} token during pretraining to stabilize multi-step reasoning.

Another line of work seeks to internalize reasoning behavior by distilling chain-of-thought rationales into latent representations. ~\cite{deng2023implicit} trains models to mimic CoT-style reasoning implicitly through hidden states, and ~\citep{deng2024explicit} further improves inference efficiency by removing explicit intermediate steps altogether. ~\cite{yu2024distilling} proposes to distill latent reasoning capabilities into a model by supervising it with data generated for complex reasoning. More recently, ~\cite{hao2024training} go further by replacing CoT tokens with continuous latent embeddings, enabling unconstrained reasoning in the latent space to explore on complex tasks including math and logical reasoning.
While prior work primarily focuses on enhancing efficiency or structural planning within the LLM's latent space, our approach takes a different perspective: we treat latent tokens as a \emph{bridge for exploring visual information} into the model.


\section{Multimodal Reasoning with Latent Visual Tokens}
\label{sec:method}

\begin{figure}[t]
\centering
\includegraphics[width=1\linewidth]{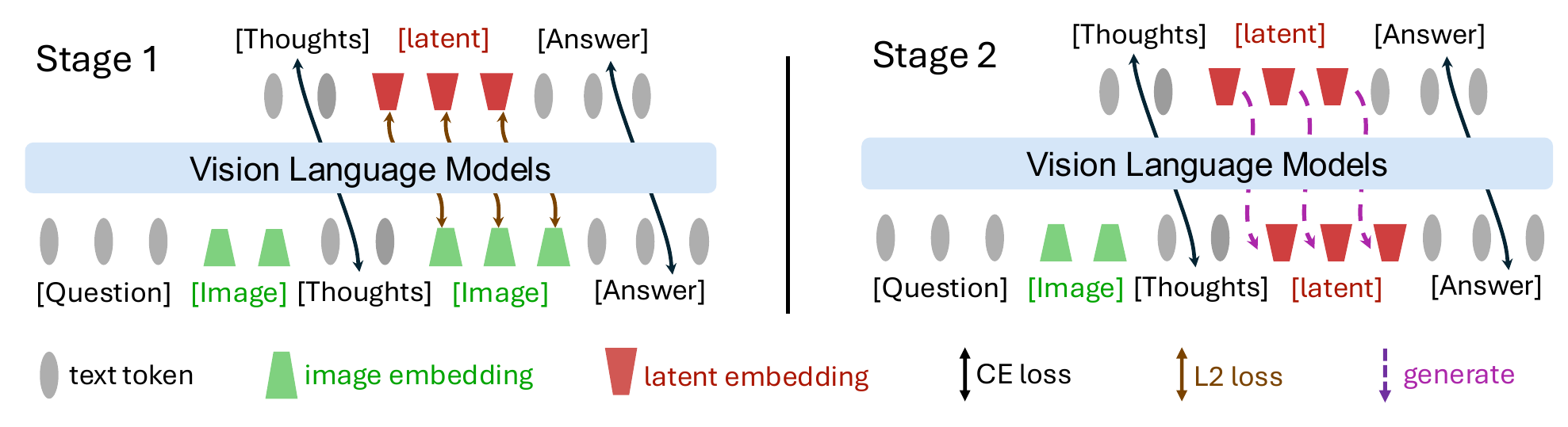}
\caption{\textbf{Pipeline of \Model Framework.} Stage 1 jointly supervises text and latent visual tokens, grounding the latter in the visual subspace; Stage 2 drops the latent supervision, anchoring the grounded latent tokens for subsequent text generation.}
\label{fig:framework}
\end{figure}

Inspired by the cognitive process of mental imagery, we introduce \Model, a framework that lets VLMs reason in interleaved multimodal trajectories. In contrast to prior unified models that integrate an external image decoder and pre-train on large-scale image generation, our method generates compact latent embeddings that serve as visual tokens. By sidestepping image generation, the model can devote its capacity to reasoning, producing only the essential visual cues and thereby echoing the concise, sketch-like representations humans employ during reasoning.

In this section, we first explain how we synthesize informative multimodal reasoning data (Sec.~\ref{sec:method-data-gen}). Next, we introduce our first joint supervision training stage in Sec.~\ref{sec:method-stage1}. Finally, we explain the second stage, which applies text-only supervision while relaxing the latent constraints (Sec.~\ref{sec:method-stage2}).

\subsection{Data Generation}
\label{sec:method-data-gen}

Consider the multimodal reasoning task where the VLMs need to generate responses $y$ to the input that consists of one or more images and a textual query. 
For simplicity, we denote the input that contains both image and text as $\bm x$.

\begin{figure}
    \centering
    \includegraphics[width=\linewidth]{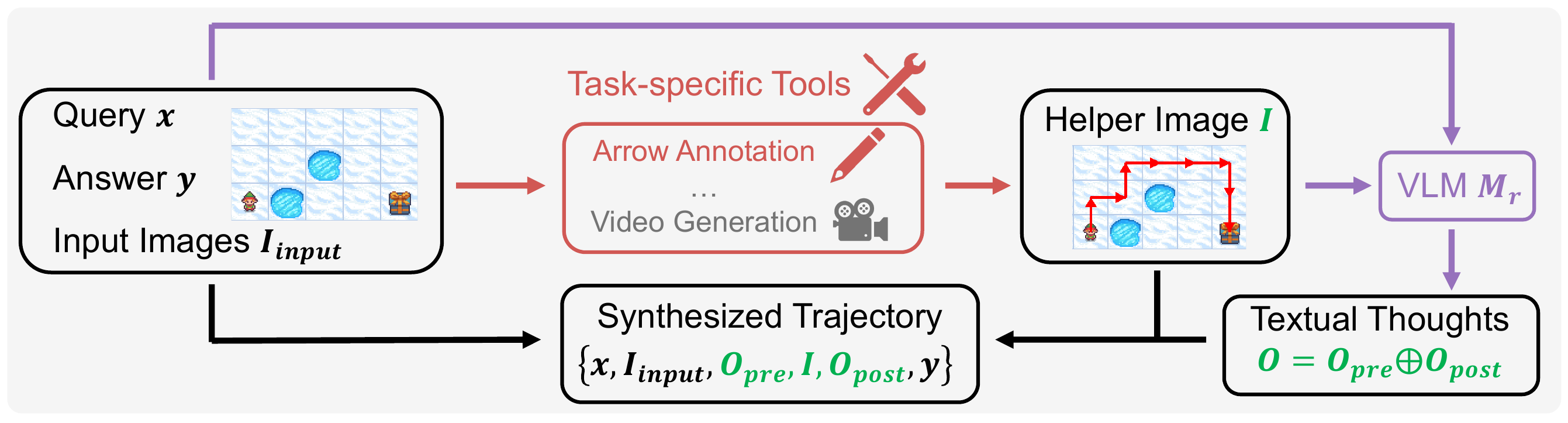}
    \caption{\textbf{Data-generation Pipeline.} For each question–answer pair, we first create a helper image with task-specific tools (here, annotate the map with arrows), then prompt a VLM to produce textual reasoning that embeds this image. The text and helper image together form the synthetic multimodal trajectory used for training.}
    \label{fig:data-example}
\end{figure}

Given VLMs naturally generating text tokens only, they require additional supervised fine-tuning to learn an interleaved reasoning pattern.
We therefore begin by synthesizing a training corpus that pairs each input $\bm x$ with a task-specific helper image $I$ (See Fig.~\ref{fig:data-example}).
For example, in the navigation task, the helper image can be generated by taking the ground truth action list and manually drawing the corresponding path on the starting map with red arrows. Similarly, for the jigsaw task, we can concatenate the candidate fragments to form a composite image that captures the relationship among pieces. More details on the image generation procedures for different tasks can be found in the supplementary materials. In general, we obtain a help image that delivers precisely the visual cues needed to supervise latent reasoning.

With the helper image $I$ prepared, we next synthesize a reasoning chain where the LLM incorporates the helper image to generate the final solution.
Specifically, we first feed a large reasoning VLM $M$ with the original input $\bm x$, the ground-truth answer $\bm y$, and the helper image $I$ and prompt it to generate a step-by-step reasoning that incorporates the helper image. For example, the prompt can be \texttt{Generate a step-by-step reasoning that leads to the ground-truth answer while properly incorporating the helper image in reasoning}. Denote the model response as
\begin{equation*}
    \bm o = M\left(\bm x, \bm y, I \right).
\end{equation*}
Here $\bm o$ is a step-by-step reasoning with the helper image embedded in the reasoning process.
Since the helper image is embedded in the reasoning chain, it splits the reasoning chain into two parts.
Without loss of generality, we represent $\bm o = \bm o_{\rm pre} \oplus I \oplus \bm o_{\rm post}$, where $\oplus$ is the concatenation operation, $\bm o_{\rm pre}$ is the reasoning chain before the helper image while $\bm o_{\rm post}$ is the reasoning chain after the helper image.
By prompt the large reasoning VLM with different inputs, we can thus collect a training dataset $\mathcal{D} = \{\bm x^{(i)}, I^{(i)}, \bm o^{(i)}, \bm y^{(i)}\}_{i=1}^{N}$, where each $\bm o^{(i)}$ is a synthesized reasoning chain with text and image interleaved.

\subsection{Joint Supervision for Latent Grounding}
\label{sec:method-stage1}

To teach the model an interleaved style of reasoning, one naive solution is to directly train a VLM on the data collected above. However, the effectiveness can be negatively affected by the model's limited capability of synthesizing helper images. Therefore, we propose a novel training strategy: pass the helper images to the VLM first to convert the helper images in the synthetic training data into patch-level features; then fine-tune the VLM to output such features as latent reasoning tokens, thus eliminating the need to generate helper images by the VLM.

More specifically, for each training example $(\bm x, I, \bm o, \bm y)\in \mathcal{D}$, we pass the helper image $I$ through the VLM $f_{\bm \theta}(\cdot)$ with parameter $\bm \theta$ to obtain its patch-level features $\{\bm e_{1},\dots,\bm e_{n}\}=f_{\bm \theta}(I)$. Rather than asking the model to reproduce every patch, we mimic human mental imagery by compressing these features into $k$ salient vectors, $\{\hat{\bm e}_{1},\dots ,\hat{\bm{e}}_{k}\}=\mathrm{Compress}\bigl(\{\bm e_{1},\dots ,\bm e_{n}\}\bigr)$, that retain only task-critical visual cues. In this work, we realize ${\rm Compress}(\cdot)$ with simple average pooling over the original patch features, a lightweight yet surprisingly effective strategy that supplies a concise visual summary for supervision. We then train our model to
(1) generate the response $\bm o_{\rm pre}$ conditioned on the input $\bm x$,
(2) generate the latent tokens $\{\hat{\bm e}_{1},\dots ,\hat{\bm{e}}_{k}\}$ conditioned on $\bm x$ and $\bm o_{\rm pre}$, where the last layer hidden states at corresponding positions will be regarded as the generated latent tokens,
and (3) generate the response $\bm o_{\rm post}$ conditioned on the proceeding content.

For the training objective for latent token generation, we adopt the cosine similarity between the last layer hidden states of the model and the target latent tokens:
\begin{equation}
\mathcal{L}_{\rm visual}\;=\;
\ell_{\cos}\Bigl(\hat{\bm{e}}_{j},\,g_{\theta}\bigl(\bm o_{\rm pre},\,\hat{\bm{e}}_{1:j-1}\bigr)\Bigr),
\end{equation}
where $g_{\theta}\bigl(\bm o_{\rm pre},\,\hat{\bm{e}}_{1:j-1}\bigr)$ denotes the model's prediction for the $j$-th latent token conditioned on the preceding context. This loss grounds the latent tokens firmly in the visual representation space.

Meanwhile, we train the surrounding textual tokens using the standard cross-entropy loss for next token prediction. For the left segment $\bm o_{\rm pre}$ the model conditions only on earlier words, whereas for the right segment $\bm o_{\rm post}$ it also attends to the $k$ compressed visual embeddings.
\begin{equation}
    \begin{aligned}
        \mathcal{L}_{\rm text} = 
        \sum_{i=1}^{|\bm o_{\rm pre}|}\ell_{\rm CE}\bigl(\bm o_{\mathrm{pre}, i},\,f_{\theta}(\bm x, \bm o_{\mathrm{pre}, <i})\bigr)
        + \sum_{i=1}^{|\bm o_{\mathrm{post}}|}\ell_{\rm CE}\bigl(\bm o_{\mathrm{post}, i},\,f_{\theta}(\bm x, \bm o_{\mathrm{pre}}, \{\hat{\bm e}_{j}\}_{1}^{k},\bm o_{\mathrm{post}, <i}\bigr).
    \end{aligned}
\end{equation}
Here $f_{\bm \theta}(\bm x)$ denotes the next token prediction probability conditioned on the input and $\{\hat{\bm e}_{j}\}_{1}^{k}$ is the set of the ground-truth latent tokens.
The overall training objective in this stage combines this term with the visual-alignment loss $\mathcal{L}_{1}=\mathcal{L}_{\rm visual}+\gamma \mathcal{L}_{\rm text}$, where the $\gamma$ is the loss coefficient, thereby anchoring the latent tokens in visual space while teaching the model to weave them naturally into its textual thoughts.

\subsection{Text-Only Supervision with Latent Relaxation}
\label{sec:method-stage2}

The first stage grounds the latent tokens by forcing the model to reconstruct the compressed image embeddings. Although effective for visual alignment, this can over-constrain the model, diverting capacity from its primary goal of answering the question correctly, degrading the reasoning performance. Therefore, in the second stage, we remove the cosine loss altogether and keep only the cross-entropy loss over text tokens.

Although the latent tokens no longer carry an explicit loss, we still anchor them so that they meaningfully guide the following thoughts. For each training instance, the model first autoregressively produces its own latent tokens $\{\bm e_i\}_{i=1}^{k}$, with
\begin{equation}
\bm e_{j}=f_{\bm \theta}\bigl(\bm x, \bm o_{\rm pre},\, \bm e_{<j} \bigr).
\end{equation}
These self-generated embeddings replace the compressed image vectors used in Stage 1 and serve as priors for the tokens that follow the image placeholder. Therefore, the training objective becomes
\begin{equation}
    \begin{aligned}
        \mathcal{L}_{\rm text} = 
        \sum_{i=1}^{|\bm o_{\rm pre}|}\ell_{\rm CE}\bigl(\bm o_{\mathrm{pre}, i},\,f_{\theta}(\bm x, \bm o_{\mathrm{pre}, <i})\bigr)
        + \sum_{i=1}^{|\bm o_{\mathrm{post}}|}\ell_{\rm CE}\bigl(\bm o_{\mathrm{post}, i},\,f_{\theta}(\bm x, \bm o_{\mathrm{pre}}, \{\bm e_{j}\}_{1}^{k},\bm o_{\mathrm{post}, <i}\bigr).
    \end{aligned}
\end{equation}
Due to the continuous property of $\{\bm e_i\}_{i=1}^{k}$, these self-generated latent tokens are fully differentiable.
Since the next token prediction of $\bm o_{\mathrm{post}}$ is a function of the latent tokens, the gradient can be propagated to these latent tokens when minimizing the above loss on the textual tokens.
This allows us to optimize the generation of the latent tokens within the learned visual subspace, acting as flexible priors that guide subsequent text generation and yield a more adaptive, task-focused reasoning.

The overall framework of our two-stage pipeline is provided in Fig.~\ref{fig:framework}. These two stages jointly endow VLMs with the ability to generate interleaved multimodal reasoning with latent visual tokens. Empirical results in Sec.~\ref{sec:exp-results} further validate the effectiveness of our latent reasoning over naive text-only decoding. 

\subsection{Reinforcement Learning}
After the two supervised fine-tuning stages, the model has already learned to reason using both interleaved text and latent tokens. Here, we further explore whether the model's performance can be improved using reinforcement learning (RL), inspired by recent long-CoT language models~\citep{xie2025teaching,shen2025satori}. Specifically, we adopt group relative policy optimization (GRPO)~\citep{shao2024deepseekmath} for RL training. For each input query in the training set, we sample multiple responses from the model. During RL, we explicitly optimize the probabilities of textual tokens while allowing gradients to flow through the latent tokens. Following LMM-R1~\citep{peng2025lmm}, we adopt two types of rewards: accuracy and format. We consider both accuracy and format rewards. For accuracy reward, we set $r_{\rm acc}( \bm o, \bm x) = 1$ if the final answer is correct, and $0$ otherwise. For the format reward, we check whether the thinking process is enclosed between ``<think>'' and ``</think>'' tags and whether the final answer format is formatted as ``\texttt{\textbackslash boxed\{\}}'' in the output response $\bm o$. If the format is correct, the reward is 0.1; otherwise, it is 0. We then use the aggregated reward for optimization.

\section{Experiments}
\label{sec:exp}

\subsection{Experimental Settings}
\label{sec:exp-settings}

\paragraph{Benchmarks.}
We evaluate our approach on four spatial reasoning benchmarks. \textbf{VSP}~\citep{wu2024vsp} measures spatial planning in a simulated maze-navigation environment. In addition to its main task, we adopt its spatial reasoning subtask, which asks the model to predict the outcome of a prescribed action sequence. We extend the original binary choice to a three-way classification. 
\textbf{BLINK-Jigsaw}~\citep{fu2024blink} systematically evaluates the capacity of multimodal large language models to extrapolate global structural and semantic information from incomplete visual inputs, thereby assessing their proficiency in reasoning about spatial organization and maintaining perceptual coherence at a fine-grained level. \textbf{SAT}~\citep{ray2024sat} evaluates both static and dynamic spatial relations.
Additionally, we include the Mathematical Geometry subset of the recent \textbf{COMT}~\citep{cheng2025comt} to assess formal spatial reasoning in mathematical contexts.
Full dataset details are provided in the supplementary material.

\paragraph{Data Synthesis.}
For each task, we sample 1k training instances for fine-tuning and 2k instances for reinforcement learning. 
COMT uniquely provides interleaved multimodal reasoning trajectories, which we directly use as both helper images and reasoning supervision. For the other benchmarks, we synthesize helper images and reasoning thoughts following the procedure outlined in Sec.~\ref{sec:method-data-gen}.
For VSP, the helper image is either the start map annotated with the red-arrow path (planning task) or the agent's current state snapshot (reasoning subtask). In Jigsaw, we concatenate one candidate patch beside the reference image. For SAT, we prompt a powerful video generation model \texttt{CogVideoX-5B}~\citep{yang2024cogvideox} to render a scene that matches the textual description. With the generated helper image, we then employ \texttt{Qwen2.5-VL 32B}~\citep{bai2025qwen2} as the external reasoning model $M_r$ to generate textual thoughts. Specifically, three distinct reasoning trajectories are generated per helper image to encourage diversity in model outputs. Full synthesis details are provided in the supplementary material.

\paragraph{Baselines.}
We compare our approach against both text-only baselines and recent unified multimodal models. First, we fine-tune the model directly using answer labels and also evaluate zero-shot reinforcement learning without any supervised warm-up. Next, using our synthetic data, we perform chain-of-thought supervised fine-tuning (CoT SFT) and then add reinforcement learning, giving a fair comparison.
In addition, we benchmark against a unified model \textbf{Anole}~\citep{chern2024anole}, training with the same multimodal supervision, and \textbf{MVoT}~\citep{li2025imagine}, which generates action and state images but does not incorporate explicit reasoning thoughts during training.

\paragraph{Implementation Details.}
In this work, unless stated otherwise, all experiments use \texttt{Qwen2.5-VL 7B} as the base model. We perform supervised fine-tuning using a batch size of 8 and a cosine learning rate scheduler with an initial learning rate of 1e-5 for both stages. The random seed is fixed at 42 to ensure reproducibility.
Reinforcement learning is implemented with the Verl framework. Unless stated otherwise, we use a latent token size of $k=4$ and a loss coefficient of $\gamma=0.1$.

\subsection{Experimental Results}
\label{sec:exp-results}

We first evaluate the effectiveness of our method on the VSP benchmark. The results are shown in  Tab.~\ref{tab:main-exp-frozenlake}. We highlight the following findings.
First, adding latent visual tokens to the reasoning process significantly improves the reasoning capability of VLMs compared to text-only baselines.
Compared to directly fine-tuning the VLM with the synthesized data, our method achieves 3\% higher accuracy on the spatial reasoning task and 11\% on the spatial planning task. Also, with our two-stage training, \Model improves the CoT SFT + GRPO, by 2\% and 7\%, respectively. This demonstrates the effectiveness of the proposed two-stage training method. 
Also, we test our method on COMT, Jigsaw, and SAT tasks and present the results in Tab.~\ref{tab:main-exp-spatial}, where we observe the consistent performance gains on both tasks, underscoring that interleaving compact visual cues consistently strengthens spatial reasoning ability. 

\begin{table}[t]
    \centering
    \caption{\textbf{Experimental Results on Visual-Spatial Planning (VSP) tasks.} }
    \scalebox{0.82}{
    \begin{tabular}{lrrrrrrrrrrrr}
        \toprule
         \multirow{2}{*}{\textbf{VSP}} & \multicolumn{5}{c}{\textbf{Spatial Reasoning}} & \multicolumn{5}{c}{\textbf{Spatial Planning}} \\
         \cmidrule(lr){2-6} \cmidrule(lr){7-11}
        & Level 3 & Level 4 & Level 5 & Level 6 & Avg. & Level 3 & Level 4 & Level 5 & Level 6 & Avg. \\
        \midrule
        Zero-Shot & 0.32 & 0.23 & 0.40 & 0.32 & 0.32 & 0.10 & 0.08 & 0.05 & 0.01 & 0.06 \\
        Direct SFT & 0.83 & 0.81 & 0.85 & 0.86 & 0.83 & 0.88 & 0.81 & 0.73 & 0.47 & 0.72 \\
        CoT SFT & 0.88 & 0.86 & 0.80 & 0.83 & 0.84 & 0.68 & 0.53 & 0.35 & 0.31 & 0.47\\
        GRPO & 0.54 & 0.49 & 0.76 & 0.67 & 0.62 & 0.42 & 0.35 & 0.26 & 0.08 & 0.28 \\
        CoT SFT + GRPO & 0.89 &0.85 & 0.84 & 0.8 & 0.85 & 0.65 & 0.58 & 0.43 & 0.38 & 0.51 \\
        \midrule
        Anole & 0.46 & 0.51 & 0.49 & 0.63 & 0.52 & 0.02 & 0.01 & 0.00 & 0.00 & 0.01 \\
        MVoT & 0.53 & 0.64 & 0.67 & 0.59 & 0.61 & 0.21 & 0.11 & 0.08 & 0.03 & 0.11 \\
        \midrule
        Ours (Direct) & 0.86 & 0.84 & \textbf{0.88} & 0.87 & 0.86 & \textbf{0.93} & \textbf{0.83} & \textbf{0.76} & \textbf{0.51} & \textbf{0.76} \\
        Ours (CoT) & 0.87 & \textbf{0.92} & 0.86 & 0.84 & 0.87 & 0.75 & 0.63 & 0.53 & 0.39 & 0.58 \\
         + w/ GRPO & \textbf{0.92} & 0.90 & 0.86 & \textbf{0.88} & \textbf{0.89} & 0.78 & 0.65 & 0.52 & 0.43 & 0.60 \\
        \bottomrule
    \end{tabular}
    }
    \label{tab:main-exp-frozenlake}
\end{table}

Additionally, we observe that unified model-based baselines such as MVoT and Anole, despite explicitly generating image tokens, perform poorly when faced with text and image interleave reasoning. After fine-tuning on the same data, they achieve only 61\% accuracy on the spatial reasoning task and 11\% on the spatial planning task. Notably, Anole struggles to even generate valid answers for the spatial planning task post fine-tuning.
Following the setup in~\cite{li2025imagine}, we construct interleaved reasoning trajectories by combining textual thoughts with simulated state images after each action step for the spatial reasoning task. While our reproduced results are lower than those reported in their paper, we attribute this discrepancy to the difference in training data. They use 6,846 samples, whereas we training with the same 1,000 samples to ensure a fair comparison. Even when compared to their reported results, our model still gains an additional 2\% improvement.
These findings further underscore the advantage of our latent design over current unified approaches.

\begin{table}[t]
    \centering
    \caption{\textbf{Experimental Results on COMT, Jigsaw, and SAT tasks.}}
    \scalebox{0.92}{
    \begin{tabular}{lrrrrrrr}
        \toprule
        \multirow{2}{*}{Method} & \multirow{2}{*}{COMT} & \multirow{2}{*}{Jigsaw} & \multicolumn{3}{c}{SAT Synthetic} & \multirow{2}{*}{SAT Real}\\
        \cmidrule(lr){4-6}
         &  &  & GoalAim & ObjM & Avg. & \\
        \midrule
        Zero-Shot & 0.63 & 0.58 & 0.50 & 0.63 & 0.57 & 0.49 \\
        Direct SFT & 0.71 & 0.87 & 0.95 & 0.95 & 0.95 & 0.67\\
        CoT SFT & 0.75 & 0.83 & 0.97 & 0.90 & 0.94 & 0.66 \\
        GRPO & - & 0.85 & 0.85 & 0.80 & 0.83 & 0.71 \\
        SFT + GRPO & - & 0.86 & 0.93 & 0.85 & 0.89 & 0.65 \\
        \midrule
        Ours & \textbf{0.77} & \textbf{0.88} & \textbf{0.98} & \textbf{0.98} & \textbf{0.98} & \textbf{0.72} \\
        \bottomrule
    \end{tabular}
    }
    \label{tab:main-exp-spatial}
\end{table}

\begin{table}[t]
    \centering
    \caption{\textbf{Experimental Results with Qwen2.5-VL 3B on COMT, Jigsaw, and SAT tasks.}}
    \scalebox{0.92}{
    \begin{tabular}{lrrrrrrr}
        \toprule
        \multirow{2}{*}{Method} & \multirow{2}{*}{COMT} & \multirow{2}{*}{Jigsaw} & \multicolumn{3}{c}{SAT Synthetic} & \multirow{2}{*}{SAT Real}\\
        \cmidrule(lr){4-6}
         &  &  & GoalAim & ObjM & Avg. & \\
        \midrule
        Zero-Shot & 0.40 & 0.45 & 0.50 & 0.38 & 0.44 & 0.51 \\
        Direct SFT & 0.67 & 0.80 & 0.82 & 0.83 & 0.83 & 0.55\\
        CoT SFT & 0.65 & 0.59 & 0.73 & 0.88 & 0.71 & 0.54 \\
        GRPO & - & 0.54 & 0.78 & 0.80 & 0.79 & 0.54 \\
        SFT + GRPO & - & 0.72 & 0.82 & 0.85 & 0.84 & 0.52 \\
        \midrule
        Ours & \textbf{0.68} & \textbf{0.85} & \textbf{0.85} & \textbf{0.93} & \textbf{0.89} & \textbf{0.64} \\
        \bottomrule
    \end{tabular}
    }
    \label{tab:main-exp-spatial-3B}
\end{table}

We notice that on VSP spatial planning task, fine-tuning with synthesized reasoning thoughts performs significantly worse than training directly on answer labels, both with and without our latent design. Two factors likely contribute to this outcome. First, as noted in prior work~\citep{li2025think}, certain visual tasks that rely heavily on perception may not benefit from explicit reasoning during fine-tuning. Second, the synthesized thoughts are generated by Qwen2.5-VL-32B; although generally sound, they are not flawless, and any imperfections propagate into the base model. Likely, in SAT, the helper images are produced by a video generation model without ground-truth annotations, which can introduce further noise to the latent prior. Despite these challenges, our latent reasoning pipeline still closes much of the performance gap, highlighting its practical robustness.

Moreover, reinforcement learning can further improve the performance of our method. As shown in Tab.~\ref{tab:main-exp-frozenlake}, by weaving latent visual tokens within the text trajectories, instead of placing them at the start, our model can naturally explore diverse sequences. After optimizing with GRPO, \Model achieves extra gains (+2\% accuracy) on VSP tasks. These results further confirm that interleaved latent cues provide informative guidance with flexible reasoning, highlighting the potential of our latent design.

\subsection{Ablation Study}
\label{sec:exp-ablation-study}

\begin{wrapfigure}{r}{0.6\textwidth}
    \centering
    \caption{\textbf{Ablation Study of Training Stages on VSP Spatial Planning task.} Both training stages work jointly to achieve better reasoning performance.}
    \begin{tabular}{lrrrrr}
        \toprule
        & \multicolumn{5}{c}{\textbf{VSP Spatial Planning}} \\
        \cmidrule(lr){2-6}
        Method & 3 & 4 & 5 & 6 & Avg. \\
        \midrule
        Ours & 0.75 & 0.63 & 0.53 & 0.39 & 0.58\\
        - w/o Stage 1 & 0.69 & 0.58 & 0.46 & 0.36 & 0.52 \\
        - w/o Stage 2 & 0.38 & 0.19 & 0.16 & 0.09 & 0.21 \\
        \bottomrule
    \end{tabular}
    \label{tab:main-ablation-training-stage}
\end{wrapfigure}

In this section, we first conduct an ablation study to evaluate the influence of the two stages of our framework. 
Tab.~\ref{tab:main-ablation-training-stage} reports the effect of removing each training phase. Training with only the first phase, which jointly supervises text and latent visual tokens, anchors the latent embeddings but leaves them constrained and lowers performance, similar to the plight of unified models.

Training with only the second stage, which relies on text loss alone while letting latent tokens evolve freely, performs slightly better the text-only baseline. Without the grounding supplied by the first stage, the latent vectors drift into regions of the multimodal embedding space that do not aid reasoning.
This outcome contrasts with findings on LLMs in Coconut~\citep{hao2024training}, where unsupervised latent vectors can benefit subsequent reasoning. The difference indicates that visual and textual subspaces in VLMs remain heterogeneous enough that a grounding phase is effective. These ablations confirm that the first stage aligns latent tokens with visual features, the second stage allows them to adapt to the task, and both steps are necessary for the final performance. We also include a deeper analysis of the generated latent embeddings in Sec.~\ref{sec:analysis-latent-behavior}.

\begin{wrapfigure}{r}{0.5\textwidth}
    \centering
    \caption{\textbf{Ablation Study of Latent Size $k$ and Loss Coefficient $\gamma$ on VSP Spatial Reasoning.} Our training pipeline remains robust and superior performance across different hyperparameters.}
    \begin{tabular}{llrrrrr}
        \toprule
        \multirow{2}{*}{$k$} & \multirow{2}{*}{$\gamma$} &\multicolumn{5}{c}{\textbf{VSP Spatial Reasoning}} \\
        \cmidrule(lr){3-7}
        & & 3 & 4 & 5 & 6 & Avg. \\
        \midrule
        2 & 0.1 & 0.85 & 0.86 & 0.89 & \textbf{0.93} & 0.86 \\
        4 & 0.1 & \textbf{0.87} & \textbf{0.92} & 0.86 & 0.84 & 0.87  \\
        6 & 0.1 & 0.85 & 0.90 & \textbf{0.91} & 0.87 & \textbf{0.88} \\
        8 & 0.1 & 0.77 & 0.77 & 0.74 & 0.70 & 0.75 \\
        4 & 0.5 & 0.84 & 0.91 & 0.84 & 0.78 & 0.84 \\
        4 & 1 & 0.77 & 0.85 & 0.85 & 0.87 & 0.83 \\
        \bottomrule
    \end{tabular}
    \label{tab:main-ablation-latent}
\end{wrapfigure}

To delve deeper into the robustness of our framework, we investigate the influence of hyperparameters: latent token size $k$ and the multimodal loss coefficient $\gamma$. 
As Tab.~\ref{tab:main-ablation-latent} shows, adjusting the loss coefficient $\gamma$ has a moderate effect. 
A larger $\gamma$ weights the latent-alignment loss less in the first stage.
When $\gamma$ approaches infinity, the first stage becomes equivalent to skipping visual supervision entirely, in other words, the second stage. This gives a poor initialization for subsequent training.
Even so, after the second stage, each $\gamma$ tested still obtains over 80\% accuracy, which attests to the overall robustness of the framework.

We observe that varying the latent size $k$ from 2 to 6 yields consistently strong performance, with $k=6$ showing a slight improvement—highlighting the resilience of our latent design. However, increasing $k$ to 8 results in a significant performance drop around 13\%, likely due to error accumulation in longer latent sequences under autoregressive non-decoding generation. These observations are consistent with prior findings that optimal latent reasoning performance in LLMs typically occurs with fewer than 6 latent tokens~\citep{hao2024training}.

\section{Analysis}

\paragraph{Generalization to Smaller Models.}
\label{sec:analysis-3B}
To further investigate the impact of our latent design, we also evaluate performance using the Qwen2.5-VL 3B model. As shown in Tab.~\ref{tab:main-exp-spatial-3B}, results are consistent with those observed on the 7B model, demonstrating clear performance gains across both tasks. Notably, compared to text-only baselines, our \Model achieves even larger improvements—5\% on the Jigsaw task and 10\% on the SAT Real task. These findings further highlight the strength of our latent design and its potential to generalize across different model scales.

\begin{figure}
    \centering
    \includegraphics[width=\linewidth]{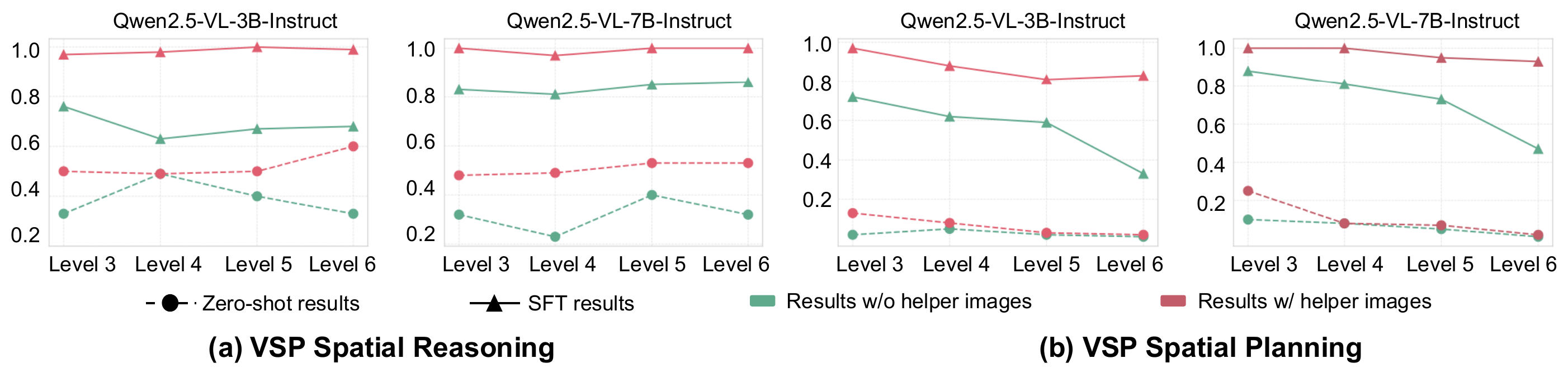}
    \caption{\textbf{Performance with Helper Images as Input Priors.} We evaluate model accuracy using synthesized helper images under both zero-shot and fine-tuned settings. The results highlight the informativeness of the generated images and confirm their high data quality.}
    \label{fig:helper-image-compare}
\end{figure}

\paragraph{Synthesized Data Quality.}
\label{sec:analysis-data-quality}
Data quality plays a critical role in model performance. In this section, we investigate whether the helper images generated by various tools are genuinely informative for VLM reasoning. For the two VSP tasks, we supply the helper image as prior input and evaluate model performance in both zero-shot and fine-tuned settings. As shown in Fig.~\ref{fig:helper-image-compare}, providing the helper image leads both models to achieve nearly 100\% accuracy on both tasks. Even in the zero-shot setting, we observe substantial performance gains on the spatial reasoning task. However, improvements on the spatial planning task are limited to simpler map layouts in the zero-shot setting. We attribute this to the inherent difficulty of extracting and leveraging spatial information from the helper image without task-specific fine-tuning.
These results suggest that the synthesized helper images do indeed enhance VLM reasoning. Moreover, if the model’s latent thoughts can fully internalize the information encoded in these images, it would represent a strong performance upper bound for our \Model.

\paragraph{Latent Behavior Analysis.}
\label{sec:analysis-latent-behavior}
During the first stage, the model learns to reproduce compressed image embeddings, anchoring its latent tokens in the visual subspace. However, after the second stage, these latent tokens receive no direct supervision. Therefore, it is unclear whether they still encode visual representations. Therefore, in this section, we further investigate the latent behaviors of our \Model.

\begin{figure}
    \centering
    \includegraphics[width=\linewidth]{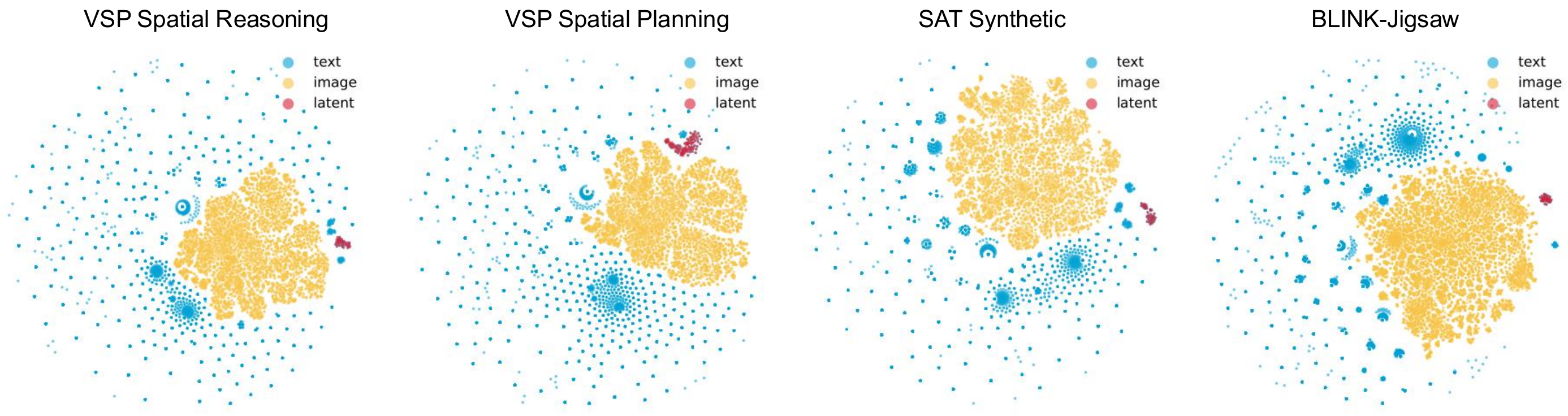}
    \caption{\textbf{Visualization of Latent Embeddings.} We visualize our latent tokens along with text and image embeddings with t-SNE. Our latent tokens cluster near, yet just outside, the visual representation subspace, consistent with the two-stage training design.}
    \label{fig:latent-viz}
\end{figure}

By sampling 100 examples from each dataset, we obtain the corresponding latent token vectors alongside the text and visual embeddings. Next, we use t‑SNE to embed all vectors into two dimensions for better visualization with a perplexity of 30, and initialize the embeddings via PCA.
As shown in Fig.~\ref{fig:latent-viz}, the text embeddings (blue dots) fill the entire plot in a radial scattering pattern, while the visual token embeddings (yellow dots) cluster tightly inside a distinct visual subspace, consistent with previous findings.
Our latent embeddings (red dots) form a compact cloud that sits just outside that visual cluster, shifted by the second training stage, which tailors the latent embeddings to answer generation.
However, we notice that our latent tokens remain clearly separated from the text distribution and closer to the visual embedding across diverse tasks.
This pattern shows that even without an explicit decoder, the latent tokens stay close to the visual manifold while retaining the flexibility introduced in the second stage, echoing the way mental imagery abstracts rather than reproduces visual input.

\section{Conclusion} \label{sec:conclusion}
In this work, mimicking human mental imagery, we propose \Model, a lightweight framework that interleaves compact latent visual tokens with text so a vision–language model can reason multimodally without ever generating pixel-level images.
Specifically, our framework is trained in two stages: a joint supervision stage that anchors latent tokens to visual embeddings while learning the surrounding text, followed by a text-only supervision stage that lets those tokens adapt freely to support answer generation.
A brief reinforcement-learning refinement further aligns the entire trajectory with task goals.
Across four spatial-reasoning benchmarks, \Model consistently outperforms text-only baselines, underscoring the effectiveness and potential of latent visual reasoning for multimodal models.

\textbf{Limitations and Future Works.} While effective, our framework has certain limitations: \textit{Synthetic Data Quality}: The performance of our interleaved reasoning depends on the quality of the generated multimodal trajectories. Carefully curating high-quality datasets for unified reasoning models is an important next step. \textit{Extend to Unified Models}: Our framework explores the latent space within a reasoning model, whereas unified models jointly align the latent space through image and text token generation during training. Despite current limitations in interleaved generation performance, whether the aligned feature space of unified models can be leveraged to further improve latent reasoning design remains an open question. \textit{Task Scale beyond Spatial Reasoning}: Currently, our evaluation is limited to spatial-reasoning benchmarks. How to extend our framework to broader multimodal or purely textual tasks remains an open direction.

\clearpage
{
    \small
    \bibliographystyle{plainnat}
    \bibliography{neurips_2025}
}

\clearpage
\appendix

\section{Datasets}

\subsection{Help Image Generation}

Diverse task-specific tools are employed to generate the helper images used in fine-tuning. In this section, we will detail the generation pipeline for each task.

\paragraph{VSP Spatial Reasoning.}
To assist in inferring the final state after a sequence of actions, we leverage the map layout visualization as the helper image, including the agent position after part of the action trajectory.
Following the VSP implementation, we render this state with the OpenAI Gym package~\citep{brockman2016openai}, using the initial map and the action sequence as inputs.

\begin{figure}[h]
    \centering
    \includegraphics[width=0.9\linewidth]{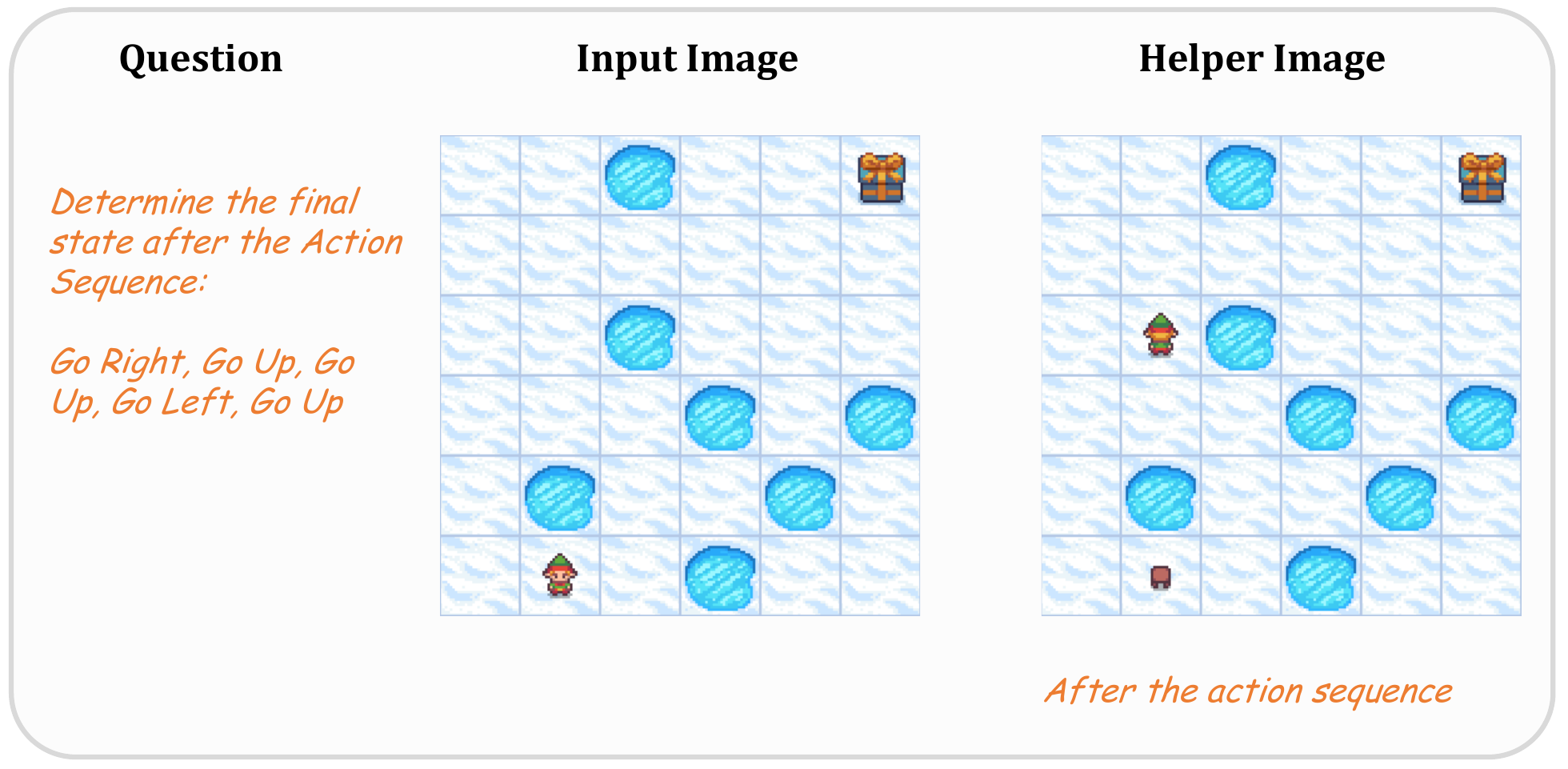}
    \caption{An example of the helper image of the VSP Spatial Reasoning task.}
    \label{fig:example-image-sim}
\end{figure}

\paragraph{VSP Spatial Planning.}
For the planning task, we provide a map annotated with the ground-truth path, turning the problem into simply reading the highlighted trajectory. Specifically, we select one valid action sequence for each sample and highlight its steps as a red arrow that begins at the agent's start position and ends at the goal.

\begin{figure}[h]
    \centering
    \includegraphics[width=0.9\linewidth]{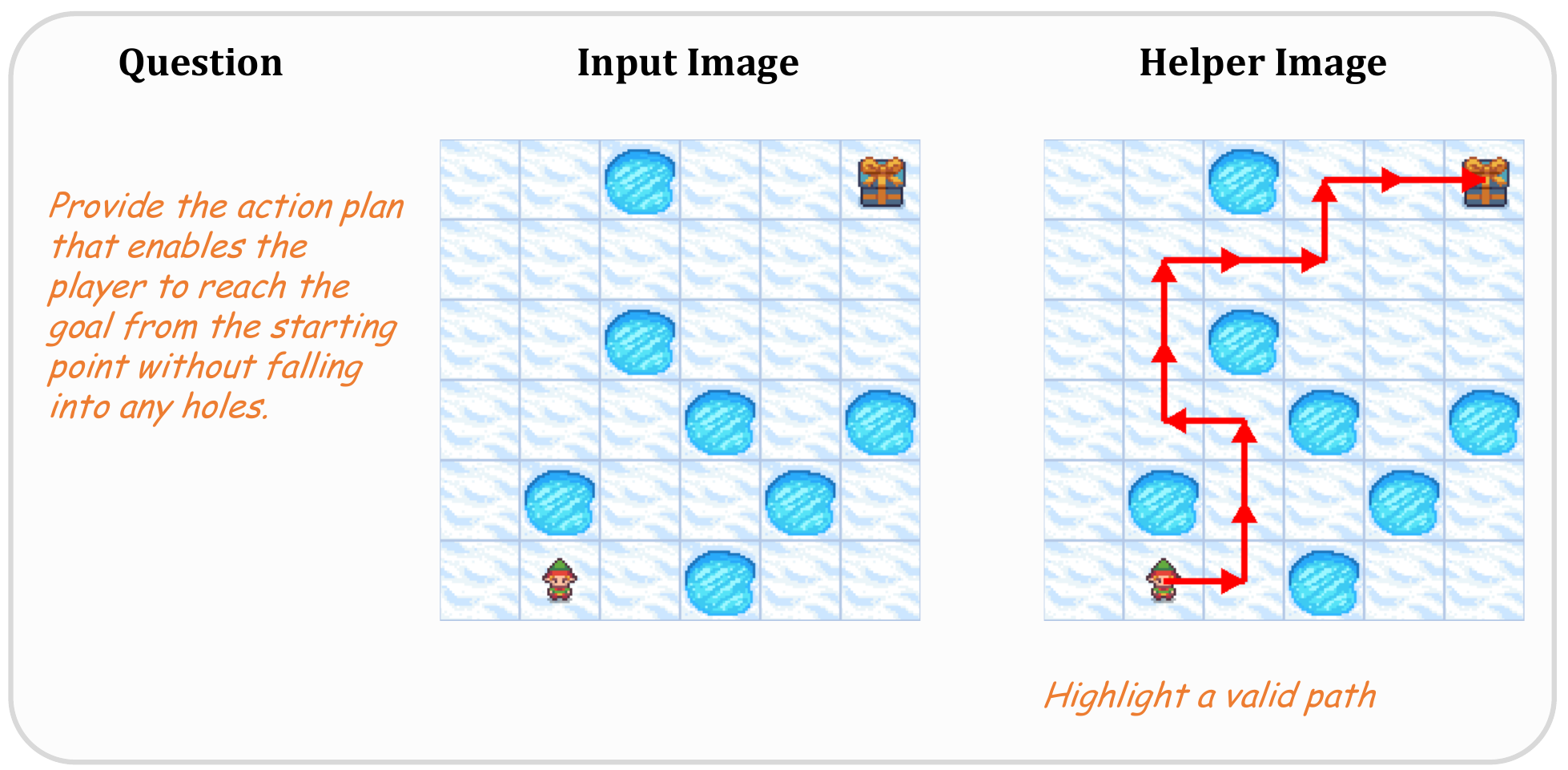}
    \caption{An example of the helper image of the VSP Spatial Planning task.}
    \label{fig:example-image-nav}
\end{figure}

\paragraph{Blink Jigsaw.}
The Jigsaw task asks which candidate patch completes the reference image. For each instance we create a helper image by inserting one randomly chosen candidate patch into the masked region. The model then can judge whether the composite looks seamless: if the patch blends smoothly, it is the correct answer; if not, the other candidate should be chosen.

\begin{figure}[h]
    \centering
    \includegraphics[width=0.9\linewidth]{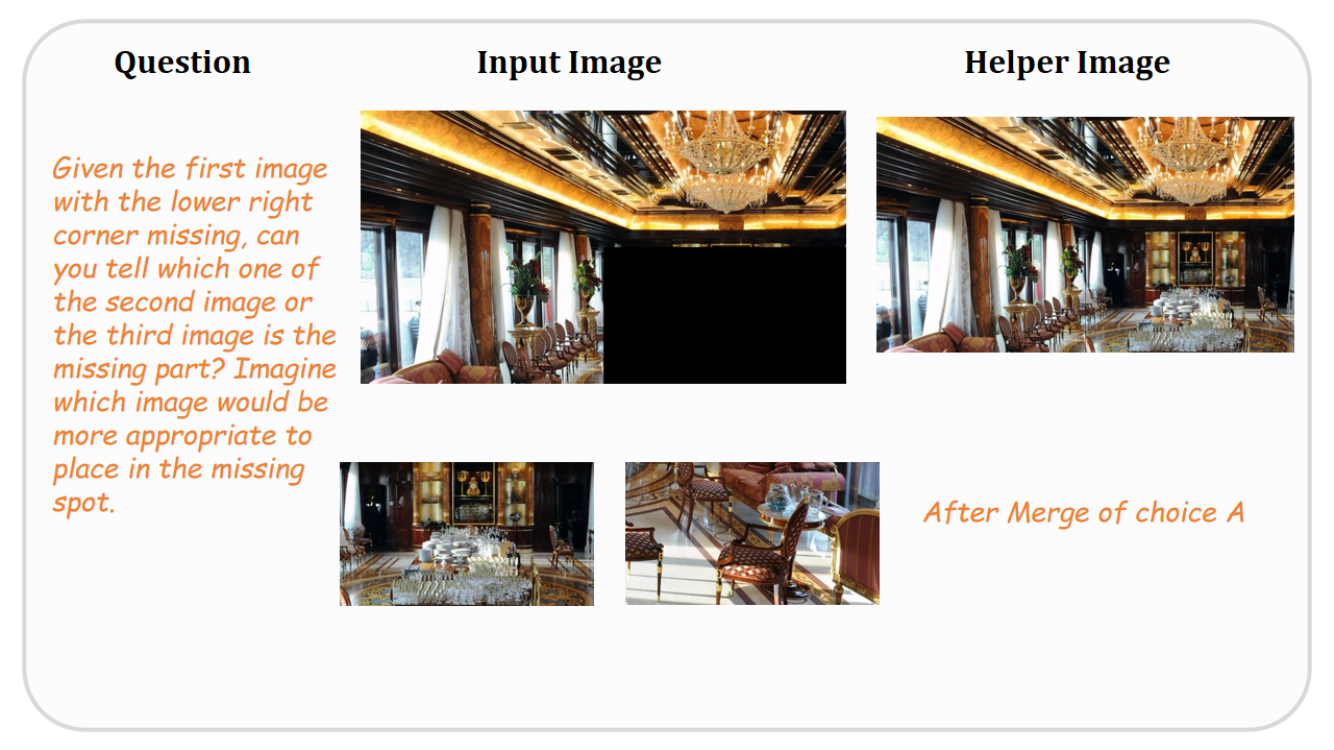}
    \caption{An example of the helper image of the BLINK task.}
    \label{fig:example-image-blink}
\end{figure}

\paragraph{SAT.}
For the SAT task, we focus on the GoalAim and ObjM subtasks, which require reasoning about a specified camera pose movement. Providing the target view as a helper image would ease the model's spatial reasoning burden. Therefore, given the recent advance in world model research, we adopt a high-quality video generation model \textbf{CogVideoX-5B} to generate this image. To further ensure the image quality, we restrict the action condition for generation to three primitives: move forward, turn left, and turn right. Sampling 9 frames along each trajectory, we instruct a VLM to choose the most informative frame. The chosen frame is then used as the helper image.

\begin{figure}[h]
    \centering
    \includegraphics[width=0.9\linewidth]{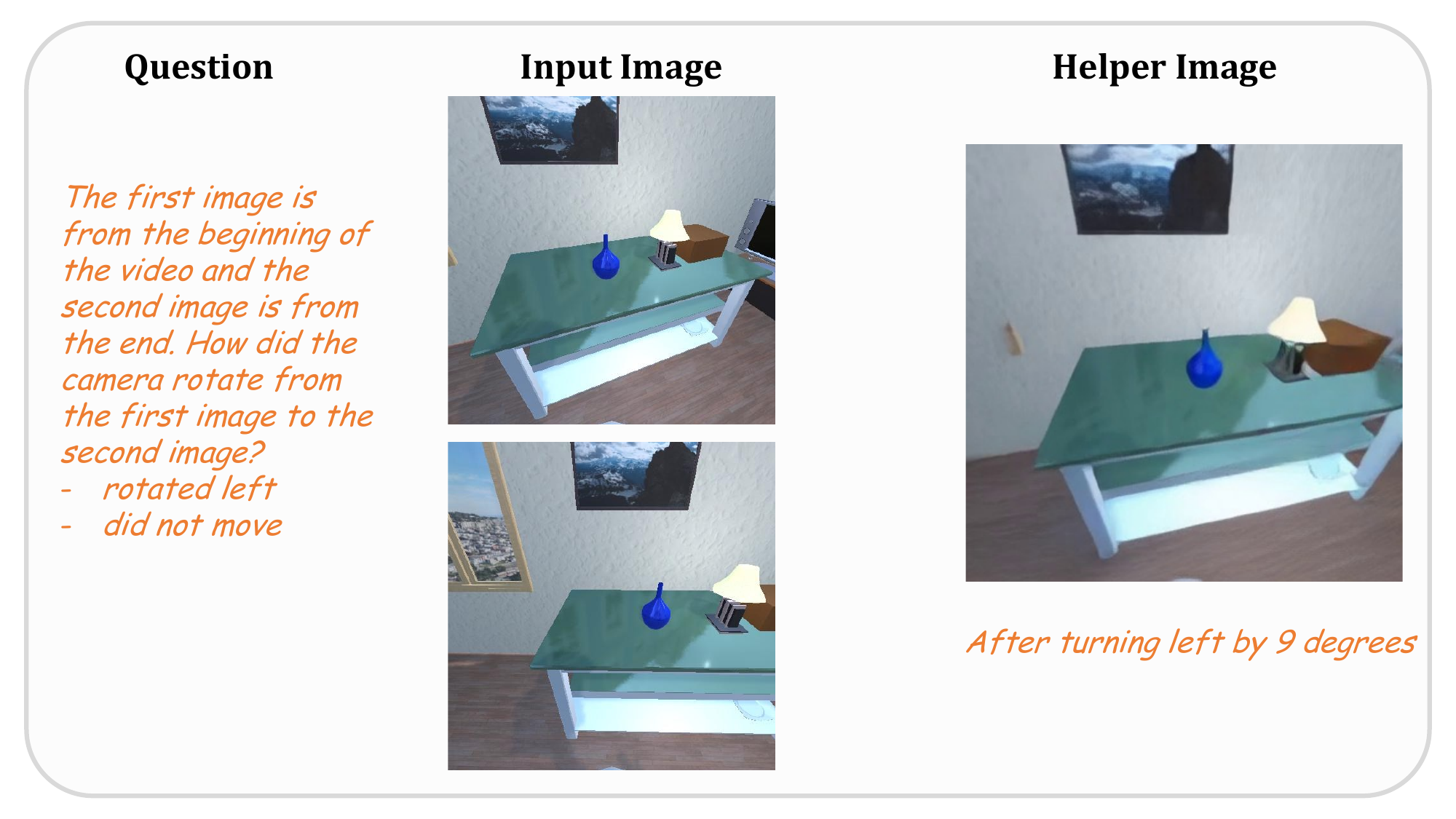}
    \caption{An example of the helper image of the SAT task.}
    \label{fig:example-image-sat}
\end{figure}

\subsection{Textual Thoughts Generation}

For each task, we generate the textual thoughts instead of leveraging closed-source outputs. We feed the helper image and the ground truth answer to a large reasoning model \texttt{Qwen2.5-VL 32B}. Task-specific prompts are applied. Simplified prompts and one illustrative example per task are provided in Tab.~\ref{tab:data-example-simulation}–\ref{tab:data-example-sat}.

The generated thoughts and the associated helper image serve as the supervision for fine-tuning, and the quality of these explanations sets an upper bound on our model's performance. 
Our current approach relies on straightforward prompts, which occasionally yield subpar reasoning trajectories.
Developing richer prompts or otherwise curating higher-quality trajectories remains an important future work.

\begin{table}[htbp]
    \centering
    \caption{Data Example of VSP Spatial Reasoning}
    \scalebox{0.96}{
    \begin{tabular}{p{\linewidth}}
        \toprule
        \makecell[c]{\textbf{VSP Spatial Reasoning}}  \\
        \midrule
        \textbf{Thoughts Generation System Prompt:}  \\
        You are analysing **one move** in FrozenLake.\\
        Tiles (numeric): 1=Start point, 0=Ice, -1=Hole, 2=Target  \\
        End-states: success, fail in hole, fail on ice\\
        \\
        Task: Given the map, current position, and one action, write **one short sentence thought** that states: 1. the square reached [(row,col), TileType]; 2. how that move shifts the likelihood of the three end-states. \\
        No bullets, no extra lines, just the single sentence. \\
        \\
        Current numeric map: \{map\_desc\}  \\
        Current position: \{position\}  \\
        Chosen action: \{action\} \\
        \\
        Example 1\\
        Numeric map: [[1,0,0],[0,-1,0],[0,0,2]]
        At (0,0) action RIGHT\\
        Thought: I slide onto [(0,1),Ice] staying safe and inching closer to the target, so success is a bit likelier while hole-failure remains unchanged.\\
        \\
        Example 2\\
        Numeric map: [[1,0,-1],[0,-1,0],[0,0,2]]\\
        At (1,0) action RIGHT\\
        Thought: Moving to [(1,1),Hole] ends the game instantly in a hole, making "fail in hole" certain and wiping out other outcomes. \\ 
        \midrule
        \textbf{Example Input:} \\
        Task: FrozenLake\\
        Given the initial state and action sequence, visualize the final state of the frozen lake map and determine whether the agent (elf character) can safely reach the gift following the action sequence without falling into the holes. If not, identify the failure reason from the given options. \\
        \\
        The definitions of the actions are as below.\\
        Go Up/Left/Down/Right: move one grid space in the absolute up/left/down/right direction.\\
        \\
        Return A, B or C.\\
        A. Action Success.\\
        B. Action Failed: Fall into the Hole.\\
        C. Action Failed: Agent Safe.\\
        \\
        Full Action Sequence: Go Left, Go Down\\
        Here is the image of the initial state of the frozenlake map. <image> \\
        \\
        The reasoning process MUST BE enclosed within <think> </think> tags. The final answer MUST BE put in $\backslash \text{boxed}\{\}$.\\
        \midrule
        \textbf{Example Output:} \\
        <think> Moving to [(1,3),Hole] ends the game instantly in a hole, making \"fail in hole\" certain and wiping out other outcomes. Next, Moving to [(1,4),Hole] ends the game instantly in a hole, making \"fail in hole\" certain and wiping out other outcomes. Next, Moving to [(2,4),Ice] keeps me on the slippery path, increasing the likelihood of success slightly while the risk of failing in a hole remains high due to nearby hazards. </think>The answer is $\backslash \text{boxed}\{\text{A}\}$.\\
        \bottomrule
    \end{tabular}
    }
    \label{tab:data-example-simulation}
\end{table}

\begin{table}[htbp]
    \centering
    \caption{Data Example of VSP Spatial Planning}
    \scalebox{0.96}{
    \begin{tabular}{p{\linewidth}}
        \toprule
        \makecell[c]{\textbf{VSP Spatial Planning}}  \\
        \midrule
        \textbf{Thoughts Generation System Prompt:}  \\
        You are a concise maze navigator. \\
        Input: (1) an image of a maze, (2) a second image, and (3) a proposed action sequence.\\
        Task: Write 1-2 English sentences that logically justify why the given action sequence reaches the treasure.\\
        - Pretend you derived the sequence yourself; NEVER say it was supplied.\\
        - NEVER mention or allude to the second image or any 'highlighted route'.\\
        - Do not output anything else.\\ 
        \midrule
        \textbf{Example Input:} \\
        As a professional maze solver, your task is to analyze a grid-based map and devise an action plan that enables a player to reach the goal from the starting point without falling into any holes, using the fewest possible moves.\\
        \\
        \#\# Game Setup \\
        - The game presents a fully observable grid-based map. \\
        - The player starts at a specified grid square, with the goal located elsewhere on the map. \\
        - Each grid square is either safe or contains a hole. \\
        - Your goal is to guide the player to the goal while avoiding holes. \\
        - For example, your final answer should be like "L,L,R,U,D", which means an action plan of left, left, right, up, and down. \\
        \\
        \#\# Moving Rules \\
        - The action plan involves a series of moves: 'L' (left), 'R' (right), 'U' (up), or 'D' (down). \\
        - Each move transfers the player to the adjacent square in that direction, provided it is a safe square. The player cannot move more than one square at a time. \\
        - Moving off the edge of the map has no effect. The player will remain at the same square. \\
        - DO NOT MOVE INTO A HOLE! Falling into a hole results in defeat. \\
        - Locating at the grid containing the goal results in victory. \\
        \\
        Here is the map: <image> \\
        Please provide your action plan. \\
        \\
        The reasoning process MUST BE enclosed within <think> </think> tags. The final answer MUST BE put in $\backslash \text{boxed}\{\}$.\\
        \midrule
        \textbf{Example Output:} \\
        <think> Moving RIGHT is the best choice because it allows me to avoid the hole directly below me and keeps me on a path that is moving closer to the goal 'G'. Moving up would lead me to a frozen square but doesn't advance me towards the goal, while moving down would immediately result in falling into a hole. Therefore, moving RIGHT is the safest and most efficient option at this moment. Next, Moving down is a strategic choice because it avoids the immediate danger of the hole to the right. It also moves the agent closer to the goal by entering a column that is likely to lead towards 'G' without any immediate threats. Moving left or right poses a higher risk of falling into a hole, whereas moving down appears to be a safe and progressive step towards the goal. Next, Moving DOWN from position is a strategic choice because it avoids the immediate danger to the left and right while still moving closer to the goal. Moving RIGHT would risk falling into the hole, and moving LEFT would take me further away from the goal. Thus, moving DOWN is the safest and most direct path towards the goal. Next, Moving DOWN is the best choice because it allows me to continue progressing towards the goal 'G' while avoiding immediate danger. To the right, there is a hole, and moving left would not bring me any closer to my objective. Moving up would be counterproductive as it would take me further away from the goal. Thus, moving DOWN is the safest and most efficient option. <image></think> The answer is $\backslash \text{boxed}\{\text{RIGHT, DOWN, DOWN, DOWN}\}$. \\
        \bottomrule
    \end{tabular}
    }
    \label{tab:data-example-navigation}
\end{table}

\begin{table}[htbp]
    \centering
    \caption{Data Example of BLINK Jigsaw}
    \scalebox{0.96}{
    \begin{tabular}{p{\linewidth}}
        \toprule
        \makecell[c]{\textbf{BLINK Jigsaw}}  \\
        \midrule
        \textbf{Thoughts Generation System Prompt:}  \\
        You are a concise visual inspector.\\
        When shown: (1) a jigsaw with a missing section, (2) two candidate pieces, and (3) the fully assembled image using the correct piece, you must give a brief justification of why the chosen piece fits. \\
        Write 1-2 English sentences, focus on visual cues such as edge continuity, color/lighting consistency, perspective, and semantics.\\  
        \midrule
        \textbf{Example Input:} \\
        Given the first image with the lower right corner missing, can you tell which one of the second image or the third image is the missing part? Imagine which image would be more appropriate to place in the missing spot. You can also carefully observe and compare the edges of the images.\\
        
        Select from the following choices\\
        (A) the second image\\
        (B) the third image\\
        
        <image><image><image>\\
        
        The reasoning process MUST BE enclosed within <think> </think> tags. The final answer MUST BE put in $\backslash \text{boxed}\{\}$.\\
        \midrule
        \textbf{Example Output:} \\
        <think> Option (B) shows a person in the foreground with a light source, which disrupts the continuity of the crowd and lighting in the main image. </think> <image> <think> The lighting and shadows in the main image suggest a more uniform and darker environment, which is not consistent with the brighter, focused light in the wrong piece. </think> $\backslash \text{boxed}\{A\}$. \\
        \bottomrule
    \end{tabular}
    }
    \label{tab:data-example-jigsaw}
\end{table}

\begin{table}[htbp]
    \centering
    \caption{Data Example of SAT}
    \scalebox{0.96}{
    \begin{tabular}{p{\linewidth}}
        \toprule
        \makecell[c]{\textbf{SAT}}  \\
        \midrule
        \textbf{Thoughts Generation System Prompt:}  \\
        You are a concise spatial-reasoning expert.\\
        Input items may include one or more scene images, a question, an additional helper image, the available answer choices, and a special field called TargetAnswer.\\
        - Use any information you need internally, but NEVER mention or allude to the additional image or the TargetAnswer.\\
        - Write 1 short English sentence that logically deduces the answer.\\
        - Embed the additional image exactly once using the literal token <image>.\\
        \midrule
        \textbf{Example Input:} \\
        You will be given one or two images and a spatial reasoning question. Your goal is to answer the spatial related question correctly. You should output an answer from the answer choice provided below.\\
        Now, according to the following image(s), answer the question from the provided choices.\\
        \\
        Question: If I turn right by 33 degrees, will I be facing away from dark frame red plaid blanket red (near the mark 2 in the image)?\\
         \\
        Answer Choice:\\
        yes\\
        no\\
         \\
        <image>\\
         \\
        The reasoning process MUST BE enclosed within <think> </think> tags. The final answer MUST BE put in $\backslash \text{boxed}\{\}$.\\
        \midrule
        \textbf{Example Output:} \\
        <image> <think> Turning right by 33 degrees will not make you face away from the bed </think> The answer is: $\backslash \text{boxed}\{no\}$. \\
        \bottomrule
    \end{tabular}
    }
    \label{tab:data-example-sat}
\end{table}

\subsection{Data Configuration}

For the Mathematical Geometry subset of COMT, we randomly sample 200 examples for evaluation and use the remaining 820 for fine-tuning.
For each other benchmark, we fine-tune on 1,000 samples and use another 2,000 non-overlapping samples for reinforcement learning. Blink-Jigsaw and SAT samples are drawn at random from their official training splits. VSP provides no training set, so we follow its data generation recipe and synthesize our own data. We ensure that each map contains a valid path, no more than 20\% trap blocks, and grid sizes 3–6 are produced in a 1:2:3:4 ratio (100, 200, 300, and 400 examples, respectively, for fine-tuning). Additionally, for each sample in VSP, we generate three distinct reasoning trajectories to encourage diversity. Full dataset statistics are provided in Tab.~\ref{tab:dataset-statistics}.

\begin{table}[h]
    \centering
    \caption{Dataset Statistics}
    \begin{tabular}{lrrr}
        \toprule
        Task & \# SFT & \# RL & \# Test\\
        \midrule
        VSP Spatial Reasoning & 3,000 & 2,000 & 400\\
        VSP Spatial Planning & 3,000 & 2,000 & 400 \\
        Blink Jigsaw & 1,000 & 2,000 & 150\\
        SAT & 1,000 & 2,000 & 500 \\
        COMT & 820 & - & 200 \\
        \bottomrule
    \end{tabular}
    \label{tab:dataset-statistics}
\end{table}


\section{Experiments}

\subsection{Implementation Details}

\paragraph{Fine-tuning.}
We adopt \texttt{Qwen2.5-VL-7B-Instruct}~\cite{bai2025qwen2} as our base VLM. The detailed training configurations are provided in Table~\ref{tab:finetune_setting}. During fine-tuning, all components of the model are trainable except for the vision encoder. The training objective combines a cross-entropy loss for next-token prediction with a cosine similarity loss for aligning latent visual tokens, as described in Sec.~\ref{sec:exp-settings}. The loss weight $\gamma$ for the visual alignment loss is set to the default value of $0.1$. Both the training stage 1 and the training stage 2 employ the same configurations.

\begin{table}[h]
\centering
\caption{
Implementation details of Supervised Fine-tuning.
}
\begin{tabular}{l|c|l|c}
\toprule
Config & Value & Config & Value \\
\midrule
optimizer & Adam & batch size &  8 \\
optimizer momentum $\beta_1 $ & 0.9 & gradient accumulation steps & 2 \\ 
optimizer momentum $\beta_2 $ & 0.95 & warmup steps & 10 \\
optimizer weight decay & 0.01 & training epochs & 10 \\
learning rate & 1e-5 & loss weight $\gamma$ & 10 \\
\bottomrule
\end{tabular}
\label{tab:finetune_setting}
\end{table}

\begin{table}[h]
\centering
\caption{
Implementation details of Reinforcement Learning.
}
\begin{tabular}{l|c|l|c}
\toprule
Config & Value & Config & Value \\
\midrule
prompt Length limit & 1024 & response length limit & 1024 \\
learning rate & 1e-6 & batch size &  32 \\
gradient accumulation steps &  4 & rollout num & 5 \\
training epochs & 15 & mini batch size & 8 \\
$\sigma_f$ & 0.1 & $\sigma_c$ & 0.9 \\
$\lambda_{kl}$ & 0.01 & $\lambda_{en}$ & 0.0 \\
\bottomrule
\end{tabular}
\label{tab:rl_setting}
\end{table}

\paragraph{Reinforcement Learning.}
We adopt \textbf{VERL}~\cite{sheng2024hybridflow} as the RL framework, and provide the detailed training settings in Tab.~\ref{tab:rl_setting}. Specifically, we utilize \textbf{Group Relative Policy Optimization (GRPO)}~\cite{shao2024deepseekmath} for reinforcement learning. The reward function consists of a \textit{format reward} and a \textit{correctness reward}, weighted by $\sigma_f$ and $\sigma_c$, respectively. KL regularization is applied with a coefficient of $\lambda_{kl}$, while entropy regularization is disabled in the policy loss by setting $\lambda_{en} = 0$. For our \Model, the KL divergence on latent visual tokens is omitted during RL training.

\subsection{Efficiency Analysis}

Both training stages of \Model are conducted on a single NVIDIA H100 GPU. Taking the VSP spatial reasoning task as an example, Stage 1 completes in approximately 3.5 hours, while Stage 2 takes around 7.2 hours. For reference, text-only CoT SFT on the same hardware requires about 5.5 hours.

\end{document}